\documentclass[conference]{IEEEtran}

\usepackage{amsmath,amsfonts,amssymb}
\usepackage{graphicx}
\usepackage{authblk}
\usepackage{url}
\usepackage{hyperref}
\usepackage{caption}
\usepackage{subfigure}
\usepackage[maxcitenames=3,natbib=true,firstinits=true,backend=bibtex,doi=false,url=false]{biblatex}
\usepackage{booktabs}
\def\BibTeX{{\rm B\kern-.05em{\sc i\kern-.025em b}\kern-.08em
    T\kern-.1667em\lower.7ex\hbox{E}\kern-.125emX}}
\usepackage{balance}

\begin{document}
\title{A Training Rate and Survival Heuristic for Inference and Robustness Evaluation (TRASHFIRE)}
\author[1]{Charles Meyers}
\author[2]{Mohammad Reza Saleh Sedghpour}
\author[1]{Tommy L\"{o}fstedt}
\author[1]{Erik Elmroth}
\affil[1]{Department of Computing Science, Ume{\aa} University, {Ume\aa}, Sweden}
\affil[2]{Elastisys AB}

\maketitle
\begin{abstract}

Machine learning models---deep neural networks in particular---have performed remarkably well on benchmark datasets across a wide variety of domains. However, the ease of finding adversarial counter-examples remains a persistent problem when training times are measured in hours or days and the time needed to find a successful adversarial counter-example is measured in seconds.
Much work has gone into generating and defending against these adversarial counter-examples, however the relative costs of attacks and defences are rarely discussed. Additionally, machine learning research is almost entirely guided by test/train metrics, but these would require billions of samples to meet industry standards. The present work addresses the problem of understanding and predicting how particular model hyper-parameters influence the performance of a model in the presence of an adversary.
The proposed approach uses survival models, worst-case examples, and a cost-aware analysis to precisely and accurately reject a particular model change during routine model training procedures rather than relying on real-world deployment, expensive formal verification methods, or accurate simulations of very complicated systems (\textit{e.g.}, digitally recreating every part of a car or a plane).
Through an evaluation of many pre-processing techniques, adversarial counter-examples, and neural network configurations, the conclusion is that deeper models do offer marginal gains in survival times compared to more shallow counterparts. However, we show that those gains are driven more by the model inference time than inherent robustness properties. Using the proposed methodology, we show that ResNet is hopelessly insecure against even the simplest of white box attacks.

\end{abstract}
\section{Introduction}

Machine Learning (ML) has become widely popular for solving complex prediction problems across many disciplines, such as medical imaging~\cite{ai_medical_imaging}, computer security~\cite{ai_security}, law enforcement~\cite{ai_prison}, aviation~\cite{ai_aviation}, and logistics~\cite{ai_luggage}. Despite this, adversarial attacks exploit ML models by introducing subtle modifications to data which leads to misclassification or otherwise erroneous outputs~\cite{chakraborty_adversarial_2018}. To ensure the robustness of ML models against adversaries has become a critical concern~\cite{adversarialpatch,carlini_towards_2017,croce_reliable_2020,hopskipjump,art2018,meyers}.

The purpose of this work was to evaluate if survival analysis can predict the success of a particular set of model hyper-parameters. In addition, we explored the relationship between computational cost and prediction accuracy in both benign and adversarial contexts.  By using samples crafted specifically to be challenging and applying survival models (see Section~\ref{aft_models}) we provide a framework to predict the expected failure time across the adversarial space. Using survival models, we demonstrate that larger machine learning models, while offering marginal gains over smaller models, do so at the expense of training times that far outpace the expected survival time and that it is simply not feasible to defend against certain attacks using the examined models and defences.

\subsection{Motivations}

It is routine to consider an adversarial context in safety---or security---critical applications~\cite{ai_medical_imaging,ai_security,ai_prison,ai_aviation,ai_luggage} where we assume the attacker is operating in their own best-case scenario~\cite{leurent2020sha,kamal2017study,madry2017towards,pixelattack,deepfool,croce_reliable_2020}. Cryptography often defines `broken' in the context of time to quantify the feasibility of an attack~\cite{leurent2020sha}---`broken' algorithms are usually defined as those for which attacks can be conducted in a (relatively) small amount of time. For example, one recent study~\cite{kamal2017study} distilled the process of password-cracking into a cloud-based service that can break common password schemes in a number of days. However, someone attacking a machine learning model might have a variety of competing goals (\textit{e.g.}, minimising the perturbation distance or maximising the false confidence)~\cite{madry2017towards,hopskipjump,pixelattack,fgm,deepfool}, so time analyses are less straightforward. What is missing, however, is a method to directly model the effect of attack criteria on the survival time.

Much work has gone into mitigating adversarial attacks, for example by adding noise in the training process~\cite{gauss_aug,gauss_out}, rejecting low-confidence results~\cite{high_conf}, or by reducing the bit-depth of the data and model weights~\cite{feature_squeezing}. However, these analyses focus on ad-hoc posterior evaluations on benchmark datasets (\textit{e.g.} CIFAR-10 or MNIST) to determine whether or not a given technique is more or less effective than another. That is, the relationship between marginal benefit and marginal cost is unclear. Furthermore, the community has trended towards larger models~\cite{desislavov2021compute} and larger datasets~\cite{desislavov2021compute,bailly2022effects} to yield increasingly marginal gains~\cite{sun2017revisiting}. For example, autonomous vehicles still largely rely on system integration tests to verify safety~\cite{vehicle_testing_review}, assuming that human-like accident metrics will guarantee safety. While there are simulation techniques~\cite{vehicle_formal} that highlight problematic scenarios by testing a component in a simulated world in which all components are modelled digitally, implementing them requires building an entire digital world that can nevertheless miss real-world edge cases. Furthermore, while formal methods for neural network verification do exist, they are generally too costly to be feasible for tuning and verifying large scale machine learning models~\cite{formal_adversarial}.
To reach safety-critical standards that are routine in other industries~\cite{iso26262,IEC61508,IEC62034}, the machine learning field must move beyond the limited test/train split paradigm that would require many, many billions of test samples for every change of a model to meet industry standards~\cite{meyers}.
The proposed method models the complex relationship between model hyper-parameters and the resulting robustness of the model, using nothing more than routine metrics collected in the model tuning stage.

\subsection{Contributions}

The contributions of this work are:
\begin{itemize}
	\item Survival analysis models for analyzing ML models under adversarial perturbations with substantial empirical evidence that survival analysis is both effective and dataset-agnostic, allowing for the expected failure rate to be predicted more precisely and accurately than with either adversarial or benign accuracy alone.
	\item Survival analysis models to measure model robustness across a wide variety of signal pre-processing techniques, exploring the relationships between latency, accuracy, and model depth.
	\item A novel metric: The \textit{training rate and survival heuristic } (TRASH) \textit{for inference and robustness evaluation} (FIRE) to evaluate whether or not a model is robust to adversarial attacks in a time- and compute-constrained context.
	\item Substantial empirical evidence that larger neural networks increase training and prediction time while adding little-to-no benefit in the presence of an adversary.
\end{itemize}

\section{Background}
Much work has gone into explaining the dangers of adversarial attacks on ML pipelines~\cite{carlini_towards_2017,croce_reliable_2020,pixelattack,fgm,biggio_evasion_2013}, though studies on adversarial robustness have generally been limited to ad-hoc and posterior evaluations against limited sets of attack and defence parameters, leading to results that are, at best, optimistic~\cite{meyers,ma2020imbalanced}. Previous work on neural network verification have relied on expensive integration tests~\cite{vehicle_testing_review}, elaborate simulation environments~\cite{vehicle_formal}, or methods that are too computationally expensive to be useful for model selection~\cite{formal_adversarial}.
However, the present work formalises methods to model the effect of attacks and defences on a given ML model and reveals a simple cost-to-performance metric to quickly discard ineffective strategies.

\subsection{Adversarial Attacks}
In the context of ML, an adversarial attack refers to deliberate and malicious attempts to manipulate the behaviour of a model. The presented work focuses on \textit{evasion attacks} that attempt to induce misclassifications at run-time~\cite{carlini_towards_2017,biggio_evasion_2013}, but note that the proposed methodology (Section~\ref{aft_models}) and cost analysis (Section~\ref{cost_normalization}) extends to other types of attacks, such as database poisoning~\cite{biggio_poisoning_2013,saha2020hidden}, model inversion~\cite{choquette2021label,li2021membership}, data stealing~\cite{orekondy2019knockoff}, or denial of service~\cite{santos2021universal}. In all sections below, metrics were collected on the benign (unperturbed) data and adversarial (perturbed) data. The abbreviations \textit{ben} and \textit{adv} are used throughout, respectively.
The strength of an attack is often measured in terms of a perturbation distance~\cite{croce_reliable_2020,chakraborty_adversarial_2018,pixelattack}. The perturbation distance, denoted by $\varepsilon\geq0$, quantifies the magnitude of the perturbation applied to a sample, $x$, when generating a new adversarial sample, $x'$. The definition is,
\begin{equation}
    \varepsilon := \| x' - x \| \leq \varepsilon^*,
    \label{eq:perturbation_distance}
\end{equation}
where $\| \cdot \|$ denotes a norm or pseudo-norm (\textit{e.g.}, the Euclidean $\ell_2$ norm or the $\ell_0$ pseudo-norm). We denote by $\varepsilon^*$ the maximum allowed perturbation of the original input. For example, this might be one bit, one pixel, or one byte, depending on the test conditions. For more information on different criteria, see Section~\ref{attacks}.

\subsubsection{Accuracy and Failure Rate}
The accuracy refers to the percentage or proportion of examples that are correctly classified. A lower accuracy indicates a higher rate of misclassifications or incorrect predictions. The accuracy, $\mathrm{Acc}$, is defined as
\begin{equation}
    \mathrm{Acc} := 1 - \frac{\mathrm{False~Classifications}}{N},
    \label{eq:acc}
\end{equation}
where $N$ is the total number of samples. The accuracy on a given test set, presumed to be drawn from the same distribution as the training set, is called the \textit{benign accuracy}, $\mathrm{Acc}_{\mathrm{ben}}$. The \textit{adversarial accuracy}, $\mathrm{Acc}_{\mathrm{adv}}$, is a measure of correct classifications in the presence of noise intended to be adversarial. However, accuracy is known to vary with things like model complexity~\cite{vgg,resnet}, data resolution~\cite{feature_squeezing}, the number of samples~\cite{vapnik1994measuring}, the number of classes~\cite{dohmatob_generalized_2019}, or the amount of noise in the data~\cite{gauss_aug,gauss_out,dohmatob_generalized_2019}. Many of these factors can also influence an attack's run time. For this reason, it is useful to think in terms of \textit{failure rate}, as
\begin{equation}
    \mathrm{Failure~Rate} := \frac{\mathrm{False~Classifications}}{\Delta t} ,
    \label{eq:failure_rate}
\end{equation}
where $\Delta t$ is a time interval. By parameterizing the measure of misclassification by time, it is possible to model the chance of failure as a function of various attributes and parameters of a model.

Let $h$ be a function that describes the rate of failure at time $t$.
This is a way to express the failure rate in terms of a \textit{hazard function}, which is defined as
\begin{equation}
    h(t) := \lim_{ \Delta t \rightarrow 0} \frac{P(t \leq T < t + \Delta t \,|\, T \geq t)}{\Delta t},
    \label{eq:failure_rate_h}
\end{equation}
where  $P(\cdot)$ is a probability and $T$ is the time until a false classification occurs, also referred to as \textit{survival time}~\cite{kleinbaum1996survival}. To be able to compare the computational efficacy of different model and attack configurations, we modelled the probability of not observing a failure before a given time, $t$, using the \textit{cumulative hazard function},
\begin{equation}
     H(t) := \int_0^{t} h(\tau) \,d\tau.
     \label{eq:cdf}
\end{equation}
Then, the \textit{cumulative survival function} is
\begin{equation}
    S(t) := P(T \geq t) = \exp(-H(t)) = 1 - F(t) = 1-   \int_0^t f(u)du
    \label{eq:S(t)}
\end{equation}
where $F(t)$ is the \textit{lifetime distribution function} which describes the cumulative probability of failure before time $t$, or $F(t) = P(T < t)$.
The probability density of observing a failure at time, $t$, is~\cite{kleinbaum1996survival},
\begin{equation*}
    f(t) := h(t)S(t).
    \label{eq:pdf}
\end{equation*}
In practice, the $h(t), S(t)$, and/or $f(t)$ can be determined whenever one of them is known~\cite{kleinbaum1996survival}.

\subsection{Cost}
\label{cost}

Assume that the cost of training a model, $C_{\mathrm{train}}$, is a function of the total training time, $T_{\mathrm{train}}$, the number of training samples, $N_{\mathrm{train}}$, and the training time per sample, $t_{\mathrm{train}} = \frac{T_{\mathrm{train}}}{N_{\mathrm{train}}}$, such that the cost of training on hardware with a fixed time-cost is
\begin{equation}
    C_{\mathrm{train}} := C_{h} \cdot T_{\mathrm{train}} = C_h \cdot t_{\mathrm{train}} \cdot N_{\mathrm{train}},
    \label{eq:naive_cost}
\end{equation}
where $C_h$ is the cost per time unit of a particular piece of hardware. Hence, the cost is assumed to scale linearly with per-sample training time and sample size, $N_{\mathrm{train}}$.
Analogously, $t_{\mathrm{predict}}$ is used elsewhere in this text to refer to the prediction time for a set of samples, divided by the number of samples. Assuming the attacker and model builder are using similar hardware then the cost to an attacker, $C_{\mathrm{attack}}$, is
\[
    C_{\mathrm{attack}} := C_{h} \cdot T_{\mathrm{attack}} = C_h \cdot t_{\mathrm{attack}} \cdot N_{\mathrm{attack}},
\]
where $ N_{\mathrm{attack}} $ is the number of attacked samples. Furthermore, a fast attack will be lower-bounded by the model inference time, $ t_{\mathrm{predict}} $, which is generally much smaller than the training time, $ t_{\mathrm{train}} $. Of course, the long-term costs of deploying a model will be related to the inference cost, but a model is clearly broken if the cost of improving a model ($\propto t_{\mathrm{train}}$) is larger than the cost of finding a counterexample ($\propto t_{\mathrm{attack}}$) within the bounds outlined in Equation~\ref{eq:perturbation_distance}. The training cost per sample does not consider how well the model performs, and a good model is one that both generalises and is reasonably cheap to train.
Therefore, a cost-normalised failure rate metric is introduced in Equation~\ref{eq:cost} in Section~\ref{cost_normalization}. Before comparing this cost to the failure rate, the attack time per sample -- or the expected survival time --  must be estimated. For that, survival models can be used.

\section{Survival Analysis for ML}
\label{aft_models}

Failure time analysis has been widely explored in other fields~\cite{aft_models}, from medicine to industrial quality control~\cite{ai_medical_imaging,ai_industry,ai_aviation,ai_luggage,ai_security,ai_prison}, but there is very little published research in the context of ML\@. However, as noted by many researchers~\cite{madry2017towards,carlini_towards_2017,croce_reliable_2020,meyers}, these models are fragile to attackers that intend to subvert the model, steal the database, or evade detection.
In this work, we leverage evasion attacks to examine the parameterised time-to-failure -- or survival time -- denoted $S_{\theta}(t)$, where $\theta$ is a set of parameters that describe the joint effect of the covariates on the survival time, usually found through maximum likelihood estimation on observed survival data~\cite{collett2023modelling}. All survival models can be expressed in terms of this parameterised survival function, $S_\theta(t)$, hazard function, $H_\theta(t)$, and lifetime probability distribution, $F_{\theta}(t)$, such that
\[
	S_{\theta}(t) := \exp\big\{-H_{\theta}(t)\big\} := 1 - F_{\theta}(t) := 1 - \int_0^t f_\theta(u) du,
\]
and the expected survival time is thus
\[
	\mathbb{E}_{S_\theta}[T] =  \int_{0}^{t^*} S_{\theta}(u) du \approx t_{\mathrm{attack}},
\]
where $t_{\mathrm{attack}}$ is an estimate of the time it takes for the average attacker to induce a failure subject to the condition in Equation~\ref{eq:perturbation_distance} and $t^*$ is the latest observed time (regardless of failure or success). The parameters, $\theta$, are estimated from model evaluation data such that: $h_{\theta}(t=t_{attack}) \approx 1 - \mathrm{Acc}_{\mathrm{adv}}$.

Survival analysis models have been widely used to investigate the likelihood of failures across fields where safety is a primary concern (\textit{e.g.}, in medicine, aviation, or automobiles)~\cite{liu2013development,lawless1995methods}. These models allow us to examine the effect of the specified covariates on the failure rate of the classifier. For manufacturing, this is done by simulating normal wear and tear on a particular component (\textit{e.g.}, a motor or aircraft sensor)~\cite{liu2013development} by exposing the component to vibration, temperatures, or impacts. For the study of diseases in humans, these models are often build on demographic data and used to examine the effect of things like age, gender, and/or treatment on the expected survival time of a patient.
Likewise, survival analysis can be used to estimate the time until a successful adversarial attack of an ML pipeline or component using metrics that are routinely collected as part of normal model training procedures. The covariates, for example, might be things like perturbation distance, model depth, number of training epochs, a signal processing technique, \textit{etc.}

These survival analysis models can broadly be separated into two categories: proportional hazard models and accelerated failure time models, each of which is outlined in the subsections below. Furthermore, by parameterizing the performance by time, it is possible to do a cost-value analysis, as outlined in Section~\ref{cost_normalization}.

\subsection{The Cox Proportional Hazard Model}
The Cox proportional hazard model tries to find model parameters, $\theta$, corresponding to covariates, $x$, to predict the hazard function on unseen configurations of the covariates, such that
\[
    h_\theta(t) = h_0(t)\phi_\theta(x) = h_0(t) \exp(\theta_1 x_1 + \theta_2 x_2 + \cdots + \theta_p x_p),
\]
where $\theta_i$ is the $i$-th model parameter and $x_i$ is the measurement of the $i$-th covariate. One downside of the Cox model compared to the accelerated failure time models discussed below is that there are few distributions that fit the proportional hazards assumption. A common choice is therefore to use a non-parametric approximation of the baseline hazards function~\cite{collett2023modelling}.
Additionally, unlike the accelerated failure time models discussed below, the coefficients in the Cox model, $\theta$, are interdependent (they are said to be \textit{adjusted} for each other) and, as such, their interpretation is not straightforward~\cite{collett2023modelling}.

\subsection{Accelerated Failure Time Models}
While Cox models assume that there is a multiplicative effect on the baseline hazard function, $h_0$, due to effect of a covariate, accelerated failure time (AFT) models instead assume that the effect of a covariate is to accelerate or decelerate the time in a baseline survival function, $S_0(t)$.
Accelerated failure time models have the form
\begin{equation} \label{eq:def_aft_model}
	S_\theta(t) = S_0 \left( \frac{t}{\phi_\theta(x)} \right),
\end{equation}
where $\phi_\theta$ is an \textit{acceleration factor} that depends on the covariates, and typically $\phi_\theta(x) =  \exp(\theta_1 x_1 + \theta_2 x_2 + \cdots + \theta_p x_p)$. The $S_0$  is the baseline survival function, and $\theta_i$ is the $i$-th acceleration factor associated with the value of the corresponding covariate, $x_i$. Common choices for parametric AFT models are listed below. Unlike the proportional hazard model discussed above, the coefficients of AFT models have a straightforward interpretation where a value of $\Phi$ represents an $\Phi$-fold increase in failure risk~\cite{collett2023modelling} and a negative value indicates a corresponding decrease in failure risk.

\subsubsection{Exponential}
The simplest AFT model of the hazard function is one that assumes a constant value over time,
\[
    h(t) = \lambda,
\]
where $\lambda$ is the false classification rate and $t$ is time. The survival time for the exponential model can be expressed as in Equation~\ref{eq:def_aft_model} when $\phi_\theta(x) = \exp \left( \theta_1 x_1 + \theta_2 x_2 + \cdots + \theta_p x_p \right) $ and
\[
    S_0(t) = \exp(-\lambda t),
\]
and if we divide $t$ by $\phi_\theta$ (letting $\phi_\theta$ consume $\lambda$), we obtain Equation~\ref{eq:def_aft_model}.

\subsubsection{Weibull}
Weibull models are flexible AFT models that assume the survival times, $T$, follow a Weibull distribution, as
\[  
    T \sim \mathcal{W}(\lambda, \sigma),
\]
where $\lambda$ and $\sigma$ are scale and shape parameters of the Weibull distribution, respectively. In the Weibull AFT model, the baseline survival function is
\[ 
    S_0(t) = \exp( {-(\lambda t)}^{\frac{1}{\sigma}}).
\]
Let
\[
    \phi_\theta(x) = \exp( \theta_1 x_1 + \theta_2 x_2 + \cdots + \theta_p x_p ).
\]
Then, the parameterised survival function can be expressed as
\[ 
    S_\theta(t) = \exp{ \left({- (\phi_\theta(x) t)}^{\frac{1}{\sigma}} \right)},
\]
letting $\phi_\theta$ consume $\lambda$.

\subsubsection{Log-Normal}
The Log-Normal AFT model assumes that the logarithm of the survival time follows a normal distribution. This model can capture scenarios where the hazard function is not monotonic over time. The logarithm of the survival time $T$ is
\[ 
   \log T \sim \mathcal{N}(\mu, \sigma^2),
\]
with mean $\mu$ and variance $\sigma^2$.
The baseline survival function is
\[ 
    S_0(t) = 1 - \Phi\left( \frac{\log t - \mu}{\sigma} \right),
\]
where $\Phi$ is the cumulative distribution function of the standard normal distribution.
Let $\mu=0$ then with
\[ 
    \phi_\theta(x) = \exp(\theta_1 x_1 + \theta_2 x_2 + \cdots + \theta_p x_p),
\]
we obtain that
\[ 
    S_{\theta}(t) = 1 - \Phi \left( \frac{\log t  -  \log(\phi_\theta(x))}{\sigma}\right).
\]


\subsubsection{Log-Logistic}
The Log-Logistic AFT model assumes that the logarithm of the survival time follows a logistic distribution. This model is useful for capturing scenarios where the hazard function first increases and then decreases over time. The survival time $T$ is expressed as
\[ 
    \log T \sim \mathcal{L}(\mu, \sigma),
\]
where $\mathcal{L}$ is a standard logistic density with mean $\mu$
and scale parameter $\sigma > 0$.  The baseline survival function is
\[ %
    S_0(t) = {\left(1 + \exp\left( \frac{\log t - \mu}{\sigma} \right)\right)} ^{-1}.
\]
Let $\mu=0$. Then, with
\[
    \phi_\theta(x) = \exp(\theta_1 x_1 + \theta_2 x_2 + \cdots + \theta_p x_p),
\]
the parameterised survival function for the Log-Logistic AFT model can be expressed as
\[ 
    S_{\theta}(t) = {\left( 1 + \exp \left( \frac{\log t - \log(\phi_\theta(x))}{\sigma}\right) \right)}^{-1}.
\]

\subsubsection{Generalised Gamma}
The Generalised Gamma AFT model is a flexible model that encompasses several other models as special cases, including the Exponential, Weibull, and Log-Normal models.
With $\phi_{\theta}(x) = \theta_1 x_1 + \theta_2 x_2 + \cdots + \theta_p x_p$, the parameterised survival function can be written as
\[
    S_{\theta}(t) = 1 - \Gamma_{\theta}\left(\rho, {\left(\lambda \frac{t}{\phi_\theta(x)}\right)}^\beta\right),
\]
where $\Gamma_{\theta}$ is the incomplete Gamma function, $\lambda > 0$ is a scale parameter, and $\rho>0$ and $\beta > 0$ are shape parameters.

\subsection{Survival Model Validation}
\label{metrics}

To compare the efficacy of different parametric AFT models, we use the Akaike information criterion (AIC) and the Bayesian information criterion (BIC)~\cite{stoica2004model,taddy2019business}, where the preferred model will be the one with the smallest value. We provide the
concordance score, which gives a value between $0$ and $1$ that quantifies the degree to which the survival time is explained by the model, where a $1$ reflects a perfect explanation~\cite{kleinbaum1996survival} and 0.5 reflects random chance. We also include graphical calibration curves (see Figure~\ref{fig:aft_models}), that depict the relationship between the fitted model (second axis) and a model fit to the data using cubic splines (first axis) as proposed by Austin \textit{et al.}~\cite{ici}. We also measured the mean difference between the predicted and observed failure probabilities,  called the integrated calibration index (ICI) as well as the error between these curves at the 50\textsuperscript{th} percentile (E50). Except for AIC and BIC, we have provided these metrics for both the training and test sets, the latter of which was 20\% of the total number of samples.

\section{Failure Rates and Cost Normalisation}
\label{cost_normalization}

With an estimate for the expected survival time, the cost-normalised failure rate, or training time to attack time ratio, can be quantified. Under the assumption that the cost scales linearly with $t_{\mathrm{train}}$ (as in Equation~\ref{eq:naive_cost}), one can divide this cost by the expected survival time to get a rough estimate of the relative costs for the model builder ($C_{\mathrm{train}} \propto t_{\mathrm{train}}$) and the attacker ($C_{\mathrm{adv.}} \propto t_{\mathrm{attack}} \approx \mathbb{E}_{S_\theta}[T]$). Recalling the definition of $\varepsilon$ in Equation~\ref{eq:perturbation_distance}, the cost of failure in adversarial terms can be expressed as,
\begin{equation}
	\bar{C}_{\mathrm{adv.}}=\frac{t_{\mathrm{train}}}{\mathbb{E}_{\theta}[T \,|\, 0 < \varepsilon \leq \varepsilon^*]}.
	\label{eq:cost}
\end{equation}
If $\bar{C} \gg 1$  then the model is \textit{broken} since it is cheaper to attack the model than it is to train it.
The numerator can be thought as the approximate training time per sample, or training rate, and the denominator is the expected survival heuristic.
The ratio of these allows one to quantify the comparative cost of the model builder and the attacker and the coefficients of the survival model provide a way to estimate the effects of the covariates.
We call this metric the TRASH score since it quickly indicates whether more training is likely to improve the adversarial robustness and any score $ > 1$ indicates that a given model is, in fact, irredeemable.
Therefore, it be immediately discarded as \textit{broken}.


\section{Methodology}
\label{methods}
Below we outline the experiments performed and the hyper-parameter configurations of the models, attacks, and defences across the various model architectures, model defences, and attacks. All experiments were conducted on Ubuntu 18.04 in a virtual machine running in a shared-host environment with one NVIDIA V100 GPU using Python 3.8.8. All configurations were tested in a grid search using \texttt{hydra}~\cite{hydra} to manage the parameters, \texttt{dvc}~\cite{dvc} to ensure reproducibility, and \texttt{optuna}~\cite{optuna} to manage the scheduling. For each attack and model configuration, the metrics outlined in Equations~\ref{eq:acc}--\ref{eq:cost} were collected, as well as the inference time, training time, and attack generation time. A grid search was conducted over datasets, models, defences, and attacks across ten permutations of the data. For visualisation, the $f_{\mathrm{ben.}}$ and $f_{\mathrm{adv.}}$ were approximated for each attack and defence combination using Equation~\ref{eq:failure_rate}, and $\bar{C}$ was approximated in the adversarial case as per Equation~\ref{eq:cost}. Additionally, we provide links to the source code repository~\footnote{\href{https://github.com/simplymathematics/deckard/tree/main/examples/pytorch}{Our Source Code}}, as well as the source for this document and archived data~\footnote{\href{https://github.com/simplymathematics/ml_afr}{ \LaTeX~source and data for this document.}}

\subsection{Dataset}
\label{dataset}

Experiments were performed on both the CIFAR100, CIFAR10~\cite{cifar}, and MNIST~\cite{mnist} datasets. The adversarial and benign accuracies were measured together with the attack generation time and the prediction time. Equations.~\ref{eq:failure_rate}~\&~\ref{eq:cost} were used to calculate the adversarial failure rate and the cost. For accuracy, see Equation~\ref{eq:acc}. For training, 80\% of the samples were used for all datasets. Of the remaining 20\%, one-hundred class-balanced samples were selected to evaluate each attack. In addition, all data were shuffled to provide 10 training and test sets for each hyper-parameter combination. Then, the data were centred and scaled (using statistics computed from the training set to avoid data leakage). This provides a straight forward interpretation of $\varepsilon$, where $\varepsilon = 1$ implies one standard-deviation of noise.

\subsection{Tested Models}
\label{models}

The Residual Neural Network (ResNet)~\cite{resnet} is a popular classification model\footnote{More than 180 thousand citations: \href{https://scholar.google.com/scholar?cites=9281510746729853742}{ResNet citations on Google Scholar.}} because of its ability to train neural networks with many layers efficiently through \textit{residual connections}.
The residual connections allow models to have hundreds of layers rather than tens of layers~\cite{resnet,vgg}. Despite the prevalence of the reference architecture, several modifications have been proposed that trade off, for instance, robustness and computational cost by varying the number of convolutional layers in the model. We tested the \textit{ResNet-18}, \textit{-34}, \textit{-51}, \textit{-101}, and \textit{-152} reference architectures, that get their names from their respective number of layers. We used the the \texttt{pytorch} framework and the Stochastic Gradient Descent minimiser with a momentum parameter of 0.9 and learning rates $\in \{10, 1, 0.1, 0.01, 0.001, 0.0001, .00001, 0.000001\}$ for epochs $\in \{ 10, 20, 30, 50, 100\}$.

\subsection{Tested Defences}
\label{defences}

In order to simulate various conditions affecting the model's efficacy, we have also tested several defences that modify the model's inputs or predictions in an attempt to reduce its susceptibility to adversarial perturbations. Just like with the attacks, we used the Adversarial Robustness Toolbox~\cite{art2018} for their convenient implementations. The evaluated defences follow.

\textit{Gauss-in} ($\ell_2$): The `Gaussian Augmentation' defence adds Gaussian noise to some proportion of the training samples. Here, we set this proportion to 50\%, allowing to simulate the effect of noise on the resulting model~\cite{gauss_aug}. Noise levels in $\{.001, .01, .1, .3, .5, 1\}$ were tested.

\textit{Conf} ($\ell_{\infty}$): The `High Confidence Thresholding' defence only returns a classification when the specified confidence threshold is reached, resulting in a failed query if a classification is less certain. This allows to simulate the effects of rejecting `adversarial' or otherwise `confusing' queries~\cite{high_conf} that fall outside the given confidence range by ignoring ambiguous results without penalty. Confidence levels in $\{.1, .5, .9, .99, .999\}$ were tested.

\textit{Gauss-out} ($\ell_2$): The `Gaussian Noise' defence, rather than adding noise to the input data, adds noise during inference~\cite{gauss_out}, allowing to reduce precision to grey- and black-box attacks without going through costly training iterations. Noise levels in $\{.001, .01, .1, .3, .5, 1\}$ were tested.

\textit{FSQ}: The `Feature Squeezing' defence changes the bit-depth of the input data to minimise the noise induced by floating-point operations. It was included here to simulate the effects of various GPU or CPU architectures, which may also vary in bit-depth~\cite{feature_squeezing}. Bit-depths in $\{2, 4, 8, 16, 32, 64\}$ were tested.

\subsection{Tested Attacks}
\label{attacks}

Several attacks using the Adversarial Robustness Toolbox~\cite{art2018} were evaluated in order to simulate attacks that vary in information and run-time requirements across distance metrics. Other researchers have noted the importance of testing against multiple types of attacks~\cite{carlini_towards_2017}. For the purposes here, \textit{attack strength} refers to the degree to which an input is modified by an attacker, as described in Section~\ref{eq:perturbation_distance}. Below is a brief description of the attacks that were tested against. One or more norms or pseudo-norms were used in each attack, as given in the parentheses next to the attack name.

\textit{FGM} ($\ell_1, \ell_2, \ell_{\infty}$): The `Fast Gradient Method' quickly generates a noisy sample, with no feasibility conditions beyond a specified step size and number of iterations~\cite{fgm}. It generates adversarial samples by using the model gradient and taking a step of length $\varepsilon$ in the direction that maximises the loss with $\varepsilon \in \{.001,.01,.03,.1,.2,.3,.5,.8,1\}$.

\textit{PGD}  ($\ell_1, \ell_2, \ell_{\infty}$): The `Projected Gradient Method' extends the FGM attack to include a projection on the $\varepsilon$-sphere, ensuring that generated samples do not fall outside of the feasible space~\cite{madry2017towards}. This method is iterative, and was restricted here to ten such iterations. The imposed feasibility conditions on the FGM attack were in $\varepsilon \in \{.001,.01,.03,.1,.2,.3,.5,.8,1\}$.

\textit{Deep} ($\ell_2$): the `Deepfool Attack'~\cite{deepfool} finds the minimal separating hyperplane between two classes and then adds a specified amount of perturbation to ensure it crosses the boundary by using an approximation of the model gradient by approximating the $n$ most likely class gradients where $n \in \{1,3,5,10\}$, speeding up computation by ignoring unlikely classes~\cite{deepfool}. This method is iterative and was restricted here to ten such iterations.

\textit{Pixel} ($\ell_{0})$: the `PixelAttack' uses a well-known multi-objective search algorithm~\cite{pixelattack}, but tries to maximise false confidence while minimising the number of perturbed pixels. This method is iterative and was restricted here to ten such iterations. For $\varepsilon$, we tested $\{
1,4,16,64,256
\}$ pixels.

\textit{Thresh} ($\ell_{\infty})$: the `Threshold' attack also uses the same multi-objective search algorithm as Pixel to optimise the attack, but tries to maximise false confidence using a penalty term on the loss function while minimising the $\ell_2$ perturbation distance. This method is iterative and was restricted here to ten such iterations. We tested penalty terms corresponding to $\{ 1,4,16,64,256\}$

\textit{HSJ} ($\ell_2$, \textit{queries}): the `HopSkipJump' attack, in contrast to the attacks above, does not need access to model gradients nor soft class labels, instead relying on an offline approximation of the gradient using the model's decision boundaries. In this case, the strength is denoted by the number of queries necessary to find an adversarial counterexample~\cite{hopskipjump}. This method is iterative and was restricted here to ten such iterations.

\subsection{Identification of ResNet Model-, Defence-, and Attack-Specific Covariates}

For each attack, the attack-specific distance metric (or pseudo-metric) outlined in Section~\ref{attacks} was identified. To compare the effect of this measure against other attacks, the values were min-max scaled so that all values fell on the interval $[0,1]$. The same scaling was done for the defences. However, while a larger value always means more (marginal) noise in the case of attacks, a larger value for the FSQ defence indicates a larger bit-depth and more floating point error. For Gauss-in and Gauss-out, a larger number does indicate more noise, but a larger number for the Conf defence indicates a larger rejection threshold for less-than-certain classifications, resulting in less low-confidence noise. Therefore, a dummy variable was introduced for each attack and defence, allowing an estimate of their effect relative to the baseline hazard. The number of epochs and the number of layers were tracked for the models, as well as the training and inference times.

\subsection{AFT Models}
The Weibull, Log-Normal, and Log-Logistic AFT models were tested using the \texttt{lifelines}~\cite{lifelines} package in Python. The metrics outlined in Section~\ref{metrics} were used for comparisons, since they are widely used in the AFT literature~\cite{aft_models}.

\section{Results and Discussion}

Through tens of thousands experiments across many signal-processing techniques (\textit{i.e.} defences), random states, learning rates, model architectures, and attack configurations, we show that model defences generally fail to outperform the undefended model in either the benign or adversarial contexts --- regardless of configuration. Also, that the adversarial failure rate gains of larger ResNet configurations are driven by response time rather than true robustness; that these gains are dwarfed by the increase in training time; and that AFT models are a powerful tool for comparing model architectures and examining the effects of covariates. In the section below, we display and discuss the results for the CIFAR100, CIFAR10, and MNIST datasets for all attacks and defences.

\label{results}
\begin{figure*}[!hbt]
\centering
\begin{subfigure}
    \centering
    \includegraphics[width=.26\textwidth]{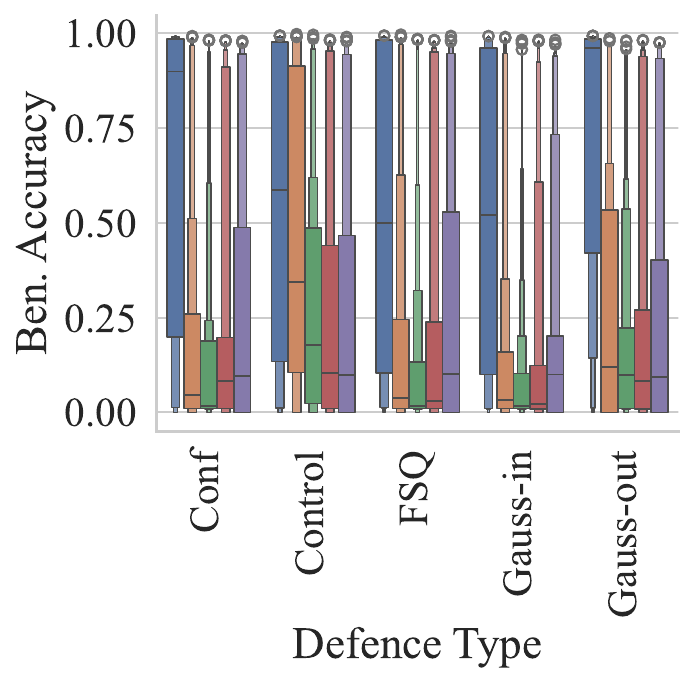}
\end{subfigure}
\begin{subfigure}
    \centering
    \includegraphics[width=.26\textwidth]{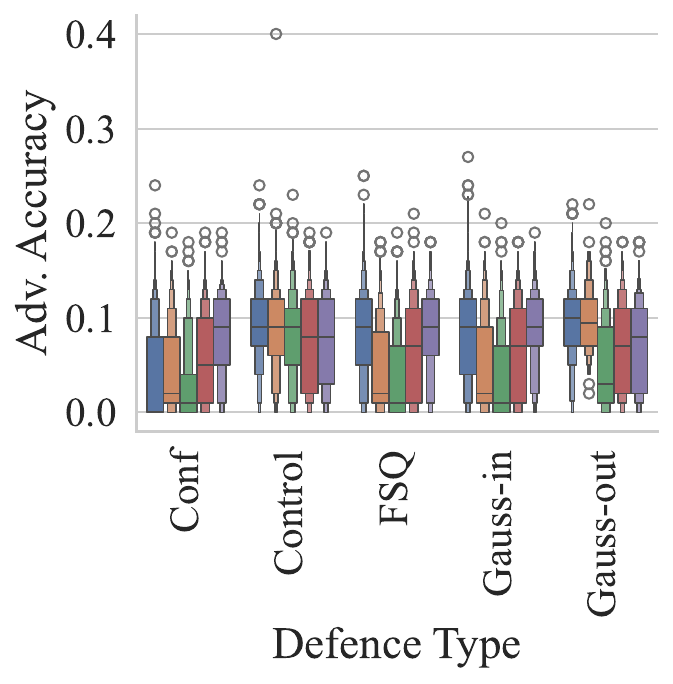}
\end{subfigure}
\begin{subfigure}
    \centering
    \includegraphics[width=.40\textwidth]{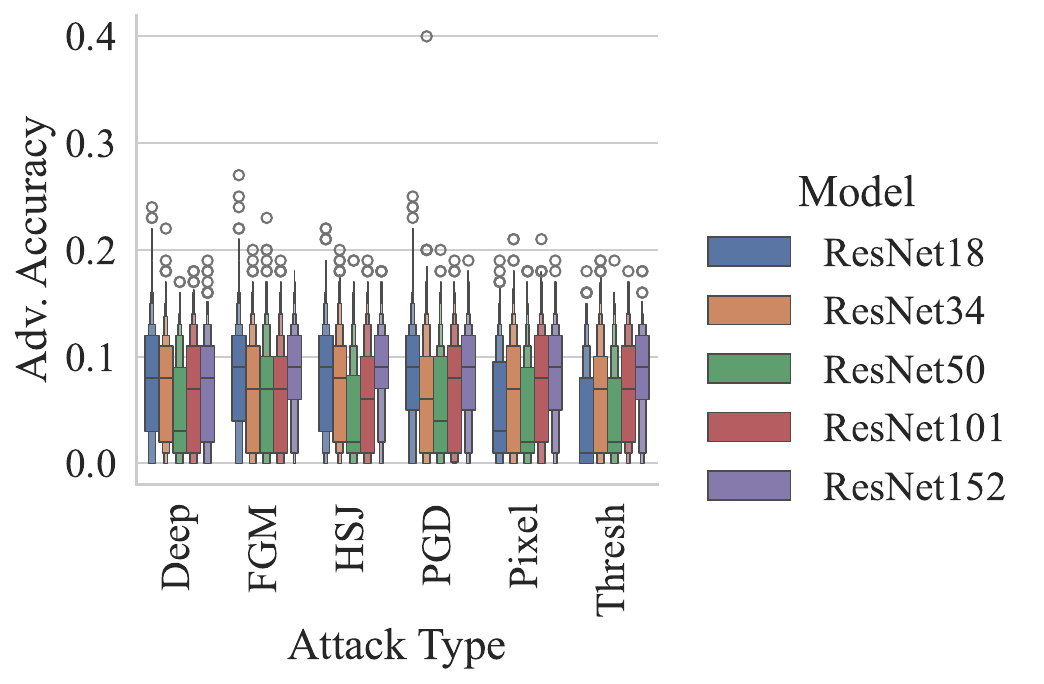}
\end{subfigure}
\caption{The adversarial accuracy across various attacks pictured on the first axis and outlined in Section~\ref{attacks}. The error bars reflect 95\% confidence intervals for the adversarial accuracy across all examined samples. The violin plots reflect 95\% confidence intervals for each tuned hyperparameter combination. Outliers are indicated with a circle.}
\label{fig:accuracies}
\end{figure*}

\subsection{Benign and Adversarial Accuracy}
Figure~\ref{fig:accuracies} depicts the benign and adversarial accuracies across the various attacks and defences. We can clearly see that no defence consistently outperforms the undefended (control) model in either the benign (left) or adversarial context (middle). We also see that for all defences, we can find at least one configuration with arbitrarily high accuracy, though finding it might involve an expensive hyperparameter search as evidenced by the large variance (see left plot of Figure~\ref{fig:accuracies}). Likewise, the adversarial accuracy benefits greatly from model tuning, but that attacks are still effective on roughly 75\% of samples in the best case for most attacks (see middle plot of Figure~\ref{fig:accuracies}). Despite varying methods and computational costs, the attacks seem to be more or less equally effective with respect to adversarial accuracy (see right plot of Figure~\ref{fig:accuracies}).

\subsection{AFT Models}

\begin{table*}[!ht]
\centering
\begin{tabular}{lllrrrrrr}
\toprule
 & AIC & BIC & Concordance & Test Concordance & ICI & Test ICI & E50 & Test E50 \\
\midrule
Cox & -- & -- & 0.92 & 0.92 & 0.07 & -- & 0.05 & -- \\
Gamma & -- & -- & 0.51 & 0.52 & 0.26 & 0.17 & 0.17 & 0.24 \\
Weibull & 9.05e+04 & 9.05e+04 & 0.92 & 0.92 & 0.02 & 0.02 & 0 & 0.01 \\
Exponential & 7.93e+04 & 7.93e+04 & 0.86 & 0.86 & 0.04 & 0.19 & 0.01 & 0.02 \\
Log Logistic & 9.79e+04 & 9.79e+04 & 0.92 & 0.92 & 0.07 & 0.08 & 0.01 & 0.01 \\
Log Normal & 1.14e+05 & 1.14e+05 & 0.91 & 0.91 & 0.15 & 0.26 & 0.08 & 0.19 \\
\bottomrule
\end{tabular}
\caption{This table depicts the performance metrics (see Section~\ref{metrics}) for various survival analysis models (see Section~\ref{aft_models}) according to the methodology described in Section~\ref{methods}.  The concordance measures the agreement between a calibration curve
(see: Section~\ref{metrics}) and the AFT model, with a value of one indicating perfect performance. The ICI score measures the total
error between the calibration curve and the fitted model and the E50 refers to the difference between the calibration curve
and fitted model at the median of either. AIC/BIC respectively refer to the Akaike and Bayesian information criteria which
favour smaller scores. Columns including the word test indicate the scores on test data, otherwise it is scored on the training set.}
\label{tab:aft_summary}
\end{table*}

Table~\ref{tab:aft_summary} contains the performance of each of these models on the CIFAR10 dataset. For all datasets, the results are roughly comparable with regards to Concordance, but the log-logistic and exponential models marginally outperforms the other models when measured with AIC/BIC\@. Concordance is identical for both the test and train sets, with gamma and exponential falling behind the others. However, the ICI  and E50  across the test train sets is superior for the Weibull, so that model was used to infer the effect of the covariates (Figure~\ref{fig:covariates}) as well as different attacks, defences, and datasets (Figure~\ref{fig:dummies}). Figure~\ref{fig:dummies} clearly shows that more hidden layers do increase the survival time. However, that seems to driven more by the model query time (see $t_{\mathrm{predict}}$ in Figure~\ref{fig:covariates}) than the number of model layers.

\begin{figure*}
	\centering
	\begin{subfigure}
		\centering
		\includegraphics[width=.25\textwidth]{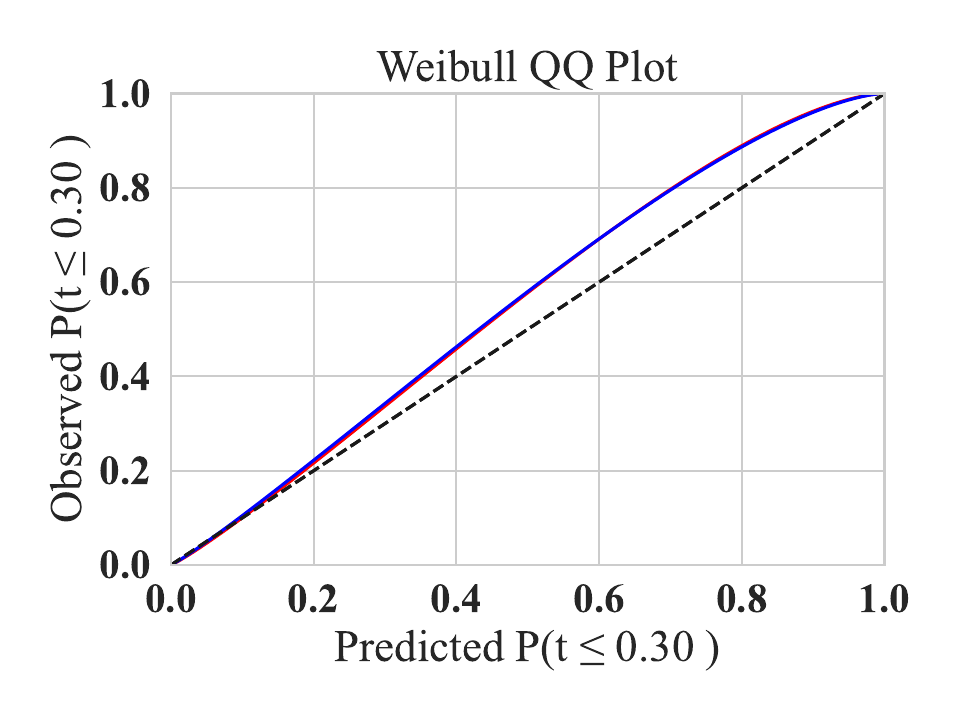}
	\end{subfigure}%
	\begin{subfigure}
		\centering
		\includegraphics[width=.25\textwidth]{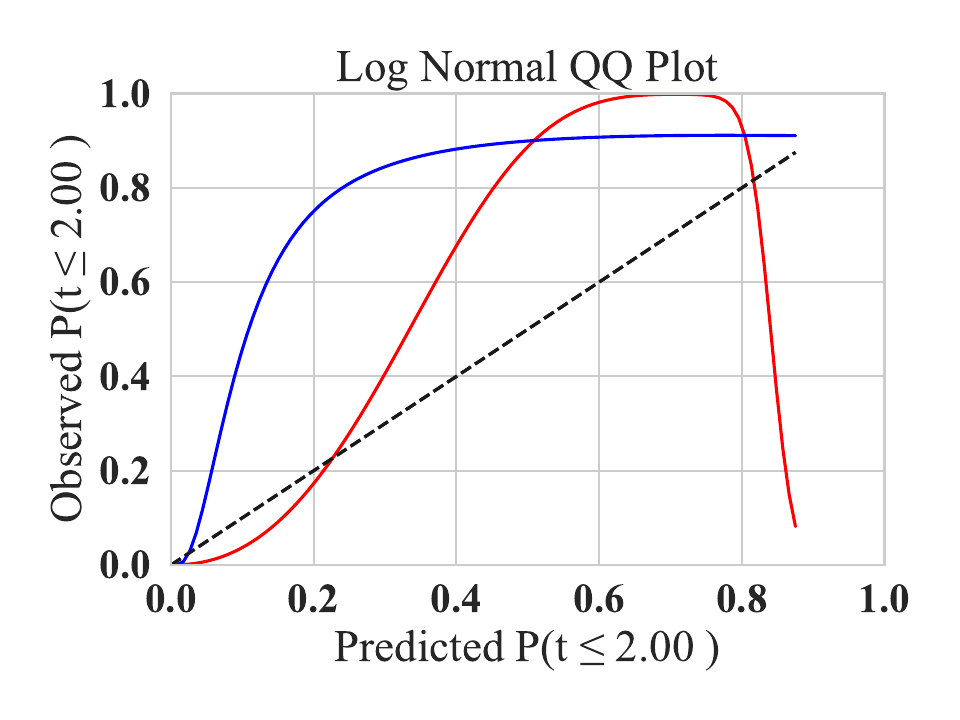}
	\end{subfigure}
	\begin{subfigure}
		\centering
		\includegraphics[width=.25\textwidth]{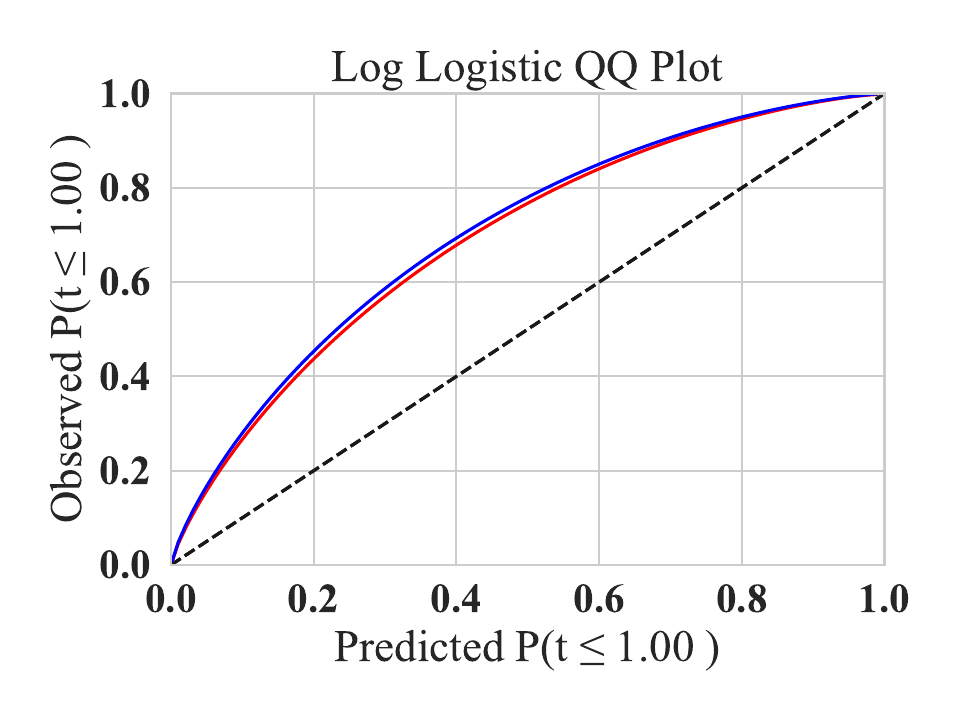}
	\end{subfigure}
 \begin{subfigure}
		\centering
		\includegraphics[width=.25\textwidth]{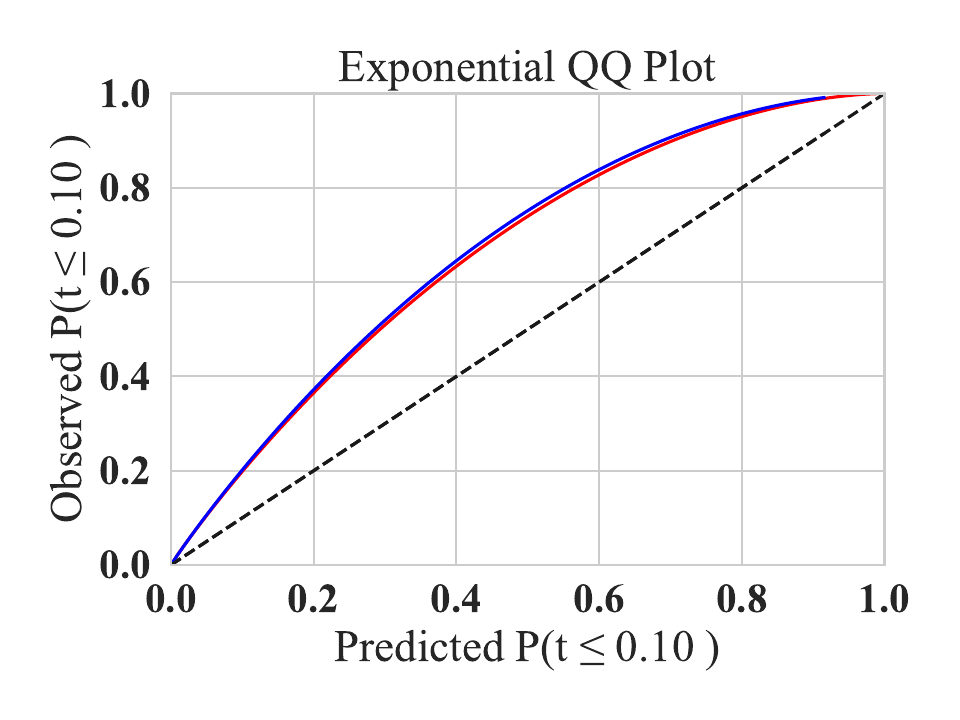}
	\end{subfigure}%
	\begin{subfigure}
		\centering
		\includegraphics[width=.25\textwidth]{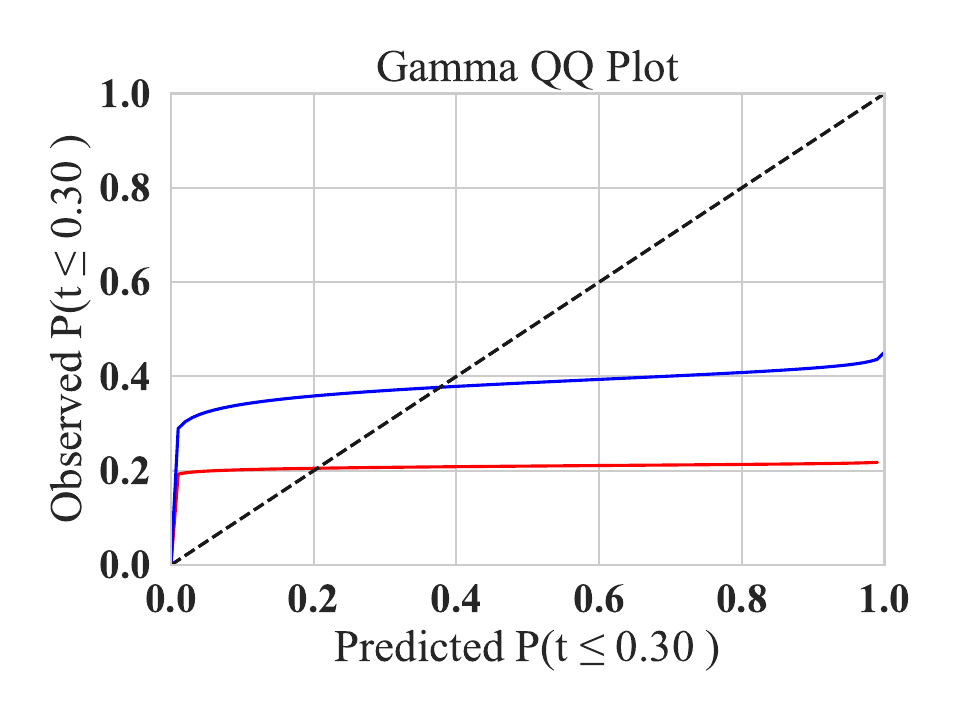}
	\end{subfigure}
	\begin{subfigure}
		\centering
		\includegraphics[width=.25\textwidth]{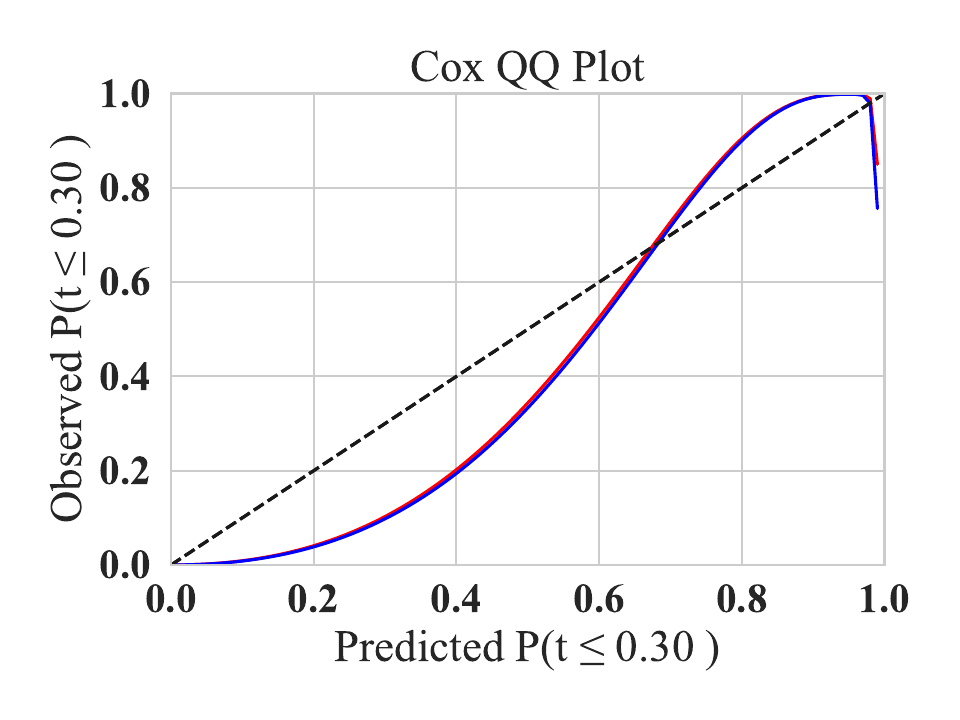}
	\end{subfigure}
	\caption{These quantile-quantile plots demonstrate the efficacy of various AFT models. The first axis is the observed quantile of a sample and the second axis represents the theoretical quantile according to the chosen AFT model. The dashed black line represents a perfect fit. To verify each model, we reserved 80\% of the data to be the training set (blue) and 20\% to be the test set (red). The time for each model was chosen to depict the best fit of the curve when the time to failure, $t$, is $ [ 0 \geq t \leq 10 ] $ seconds.
 }
	\label{fig:aft_models}
\end{figure*}

\subsection{Effect of Covariates}
\begin{figure*}
    \centering
	\begin{subfigure}
	\centering
    \includegraphics[width=.35\textwidth]{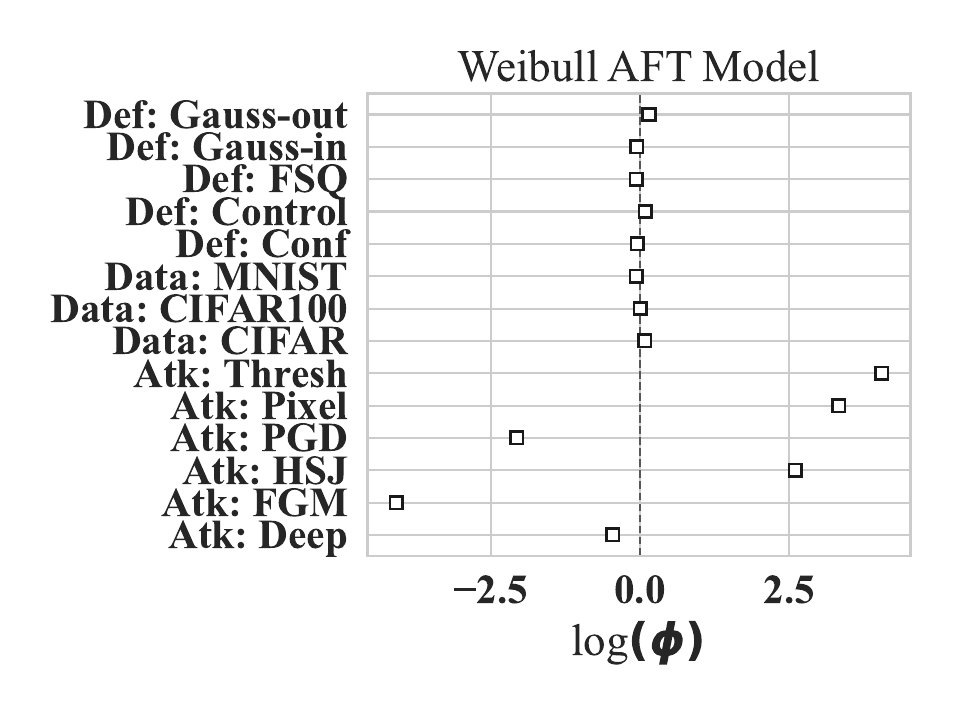}
    \end{subfigure}
    \begin{subfigure}
	\centering
    \includegraphics[width=.35\textwidth]{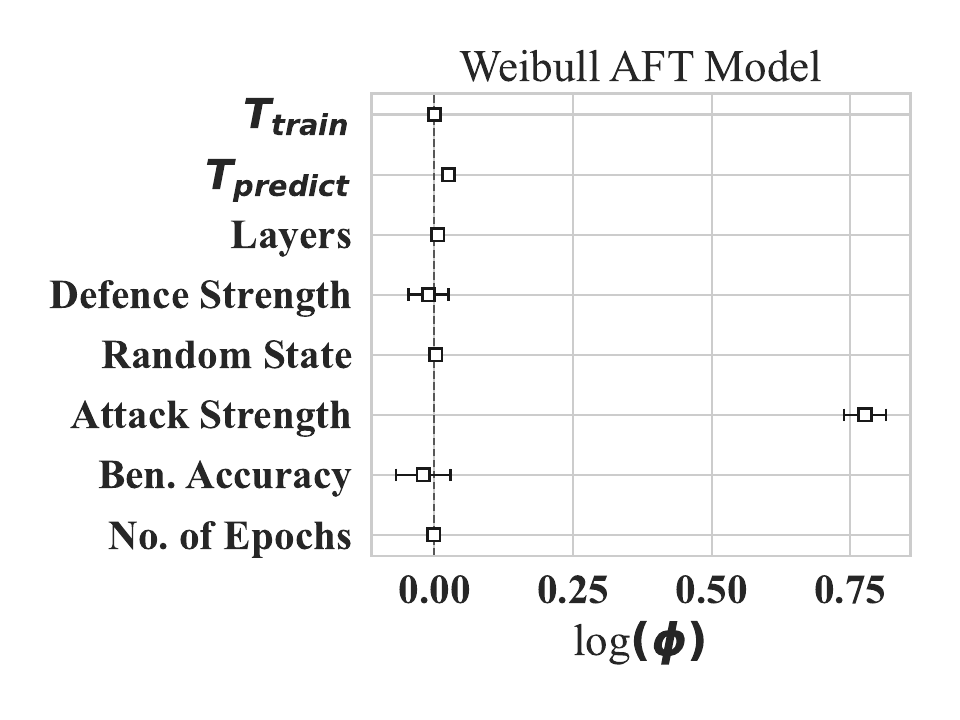}
    \end{subfigure}
    \caption{The coefficients represent the log scale effect of the dummy variables for dataset (Data), attack (Atk), and defence (Def) on the survival time, with a positive value indicating an increase in the survival time. The right plot depicts the covariates and the left plot depicts the dummy variables for the different attacks, defences, and datasets.}
    \label{fig:covariates}
    \label{fig:dummies}
\end{figure*}

Figure~\ref{fig:dummies} depicts the effect of all attacks, defences, and model configurations on the survival time and Figure~\ref{fig:covariates} depicts the effect of the covariates.
Figure~\ref{fig:covariates} clearly demonstrates that increasing the depth of the model architecture does little for adversarial robustness while universally increasing the training time.
Furthermore, it reveals something surprising -- that increasing the number of epochs tends to increase the failure rate -- even across model architectures and all defences.
Certain defences can outperform the control model -- at the cost of expensive tuning -- evidenced by the large variance in performance (see Figure~\ref{fig:dummies}). The scale of $t_{\mathrm{train}}$  in Figure~\ref{fig:covariates} shows that there is no general relationship between training time and adversarial survival time. Additionally, we see that an increase in accuracy tends to correspond to a decrease in survival time, confirming the inverse relationship noted by many researchers~\cite{carlini_towards_2017,biggio_evasion_2013,croce_reliable_2020}.
As the training time increases, however, the variance of attack times decreases, likely due to the corresponding increase in inference time (see Figure~\ref{fig:covariates}, covariate $t_{\mathrm{predict}}$) rather than inherent robustness (see covariate `Layers').
We formalise this analysis in the next subsection.

\subsection{Failures and Cost}

\begin{figure*}[!ht]
    \centering
    \begin{subfigure}
        \centering
        \includegraphics[width=.32\textwidth]{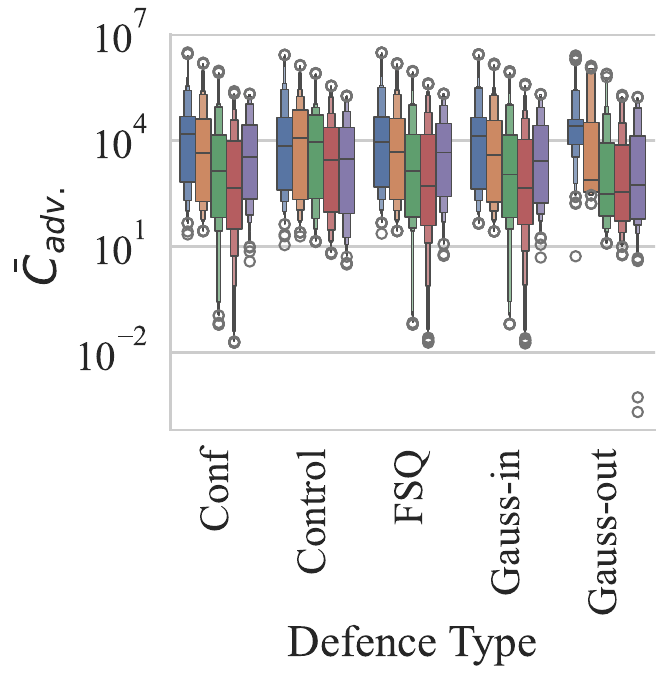}
    \end{subfigure}
    \begin{subfigure}
        \centering
        \includegraphics[width=.50\textwidth]{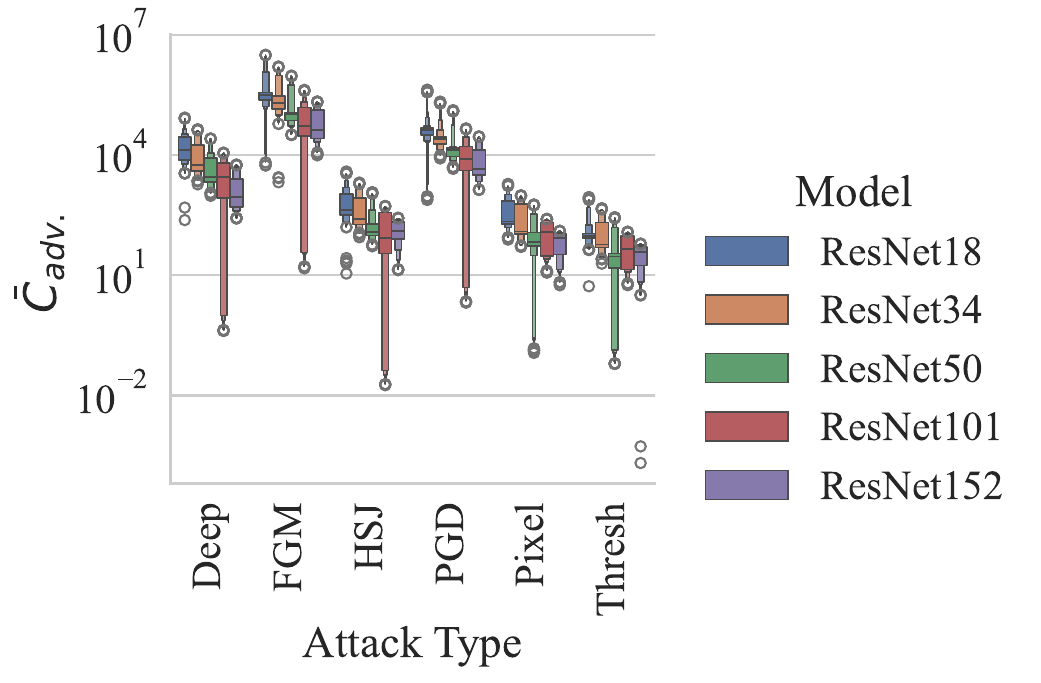}
    \end{subfigure}
    \caption{This figure depicts the TRASH metric that reflects the ratio of training-to-attack times, where a value $\gg 1 $  indicates an essential advantage for the attacker. The violin plots reflect the 95\% confidence intervals for each tuned hyperparameter combination. Outliers are indicated with a circle.}
	\label{fig:failures_per_train_time}
\end{figure*}

Figure~\ref{fig:failures_per_train_time} depicts the cost-failure ratio (see Equation~\ref{eq:cost}) in both the benign (left) figure and adversarial cases (middle and right figures), using the Weibull model to calculate $\mathbb{E}[T]$. Counter-intuitively, we see that the smallest model (ResNet18) tends to outperform both larger models (ResNet50 and ResNet152). Furthermore, we see that defence tuning is about as important as choosing the right type of defence (see left side of Figure~\ref{fig:failures_per_train_time}), with all defences falling within the normal ranges of each other. However, adding noise to the model output (Gauss-out) tends to underperform relative to the control for all models (see left side of Figure~\ref{fig:failures_per_train_time}). Likewise, the efficacy of a defence depends as much on model architecture as it does on hyperparameter tuning as demonstrated by the large variance in Figures~\ref{fig:accuracies}~\&~\ref{fig:failures_per_train_time}.  Furthermore, performance across all attacks is remarkably consistent with intra-class variation being smaller than inter-class variation almost universally across defences and model configurations.
Finally -- and most importantly -- we see that every single tested configuration performs incredibly poorly against FGM and PGD\@.

\section{Considerations}
The proposed survival and cost analysis  has some limitations that we have taken all efforts to minimise and/or mitigate.
In order to minimise timing jitter, we measured the process time for a batch of samples and then assumed that the time per sample was the measured processor time divided by the number of samples.
In order to examine a variety of different optimisation criteria for adversarial perturbations, we included several different attacks (see Section~\ref{attacks}) --- though the choice of attack is highly contextual.
We must also note that none of these attacks are run-time optimal and are, at best, an underestimate of the true adversarial failure rate~\cite{meyers}.
Likewise, testing all known defences would be computationally infeasible. So, in order to maximise the number of evaluations for each attack/defence combination, we focused only on the pre- and post-processing technique.
Techniques like adversarial retraining~\cite{croce_reliable_2020}, model transformation~\cite{papernot_distillation_2016}, and model regularisation~\cite{jakubovitz2018improving} were excluded due to their comparatively larger run-times. The equation Section~\ref{cost_normalization} reveals why techniques that significantly increase the training time might ultimately work against the model builder.
Even if one assumes there is a defence that has 99\% efficacy, rather than the, at best, 40\% efficacy indicated by the adversarial accuracy in Figure~\ref{fig:accuracies}, it would only reduce $\bar{C}_{\mathrm{adv}}$ by roughly two orders of magnitude.
While this would would be sufficient to meet the bare minimum requirement outlined in Equation~\ref{eq:cost}, in practice, attacks are often successful on even a small number of samples, whereas training often requires many orders of magnitude more. This raises serious concerns about the efficacy of any of these models and defences in the presence of these simple adversaries.
Furthermore, state of-the art leaderboards~\footnote{\href{https://github.com/MadryLab/mnist\_challenge}{Madry's MNIST Challenge}}~\footnote{\href{https://ml.cs.tsinghua.edu.cn/adv\-bench/}{Croce's Robust Bench}}
show that a 99\% generalised adversarial accuracy is, at best, optimistic. Nevertheless, the goal of this work was not to produce a comprehensive evaluation of all known defences, but to develop a cost-aware framework for evaluating their efficacy against a set of adversaries.

\section{Conclusion}
Convolutional neural networks have shown to be widely applicable to a large number of fields when large amounts of labelled data are available.
By examining the role of the attacks, defences, and model depth in the context of adversarial failure rate, this paper presents a reliable and effective modelling framework that applies AFT models to deep neural networks.
The metrics outlined Table~\ref{tab:aft_summary} and explained in Section~\ref{metrics} show that this method is both effective and data-agnostic.  We use this model to demonstrate the efficacy of various attack- and defence-tuning (see Figure~\ref{fig:aft_models}) techniques, to  explore the relationships between accuracy and adversarial robustness (Figure~\ref{fig:dummies}), and show that various model defences are ineffective on average and marginally better than the control at best.
By measuring the cost-normalised failure rate or TRASH score (see Section~\ref{cost_normalization} and Figure~\ref{fig:failures_per_train_time}), it is clear that robustness gains from deeper networks is driven by model latency more than inherent robustness (Figure~\ref{fig:aft_models}).
The methods can easily extend to any other arbitrary collection of model pre-processing, training, tuning, attack and/or deployment parameters. In short, AFTs provide a rigorous way to compare not only the relative robustness of a model, but of its cost effectiveness in response to an attacker.
The measurements rigorously demonstrate  that the depth of a ResNet architecture does little to guarantee robustness while the community trends towards larger models~\cite{desislavov2021compute}, larger datasets~\cite{desislavov2021compute,bailly2022effects}, and increasingly marginal gains~\cite{sun2017revisiting}. This approach has two advantages over the traditional train-test split method.
First, it can be used to quantify the effects of covariates such as model depth or noise distance to compare the effect of model changes. Secondly, the train-test split methodology relies on an ever-larger number of samples to increase precision, whereas the survival time method is able to precisely and accurately compare models using only a small number of samples~\cite{schmoor2000sample,lachin1981introduction} relative to the many billions of samples required of the train/test split methodology and safety-critical standards~\cite{iso26262,IEC61508,IEC62034,meyers}.
In short, by generating worst-case examples (\textit{e.g.}, adversarial ones), one can test and compare arbitrarily complex models \textit{before} they leave the lab, drive a car, predict the presence of cancer, or pilot a drone.

\printbibliography
\end{document}